\documentclass[sigconf,screen]{acmart}

\AtBeginDocument{%
  \providecommand\BibTeX{{%
    \normalfont B\kern-0.5em{\scshape i\kern-0.25em b}\kern-0.8em\TeX}}}

\copyrightyear{2021}
\acmYear{2021}
\setcopyright{acmcopyright}
\acmConference[COMPASS '21]{ACM SIGCAS Conference on Computing and Sustainable Societies (COMPASS)}{June 28-July 2, 2021}{Virtual Event, Australia}
\acmBooktitle{ACM SIGCAS Conference on Computing and Sustainable Societies (COMPASS) (COMPASS '21), June 28-July 2, 2021, Virtual Event, Australia}
\acmPrice{15.00}
\acmDOI{10.1145/3460112.3471952}
\acmISBN{978-1-4503-8453-7/21/06}

\usepackage{multirow}

\usepackage{mathtools}
\DeclareMathOperator{\boldx}{\mathbf{X}}
\DeclareMathOperator{\boldy}{\mathbf{Y}}
\DeclarePairedDelimiterX{\infdivx}[2]{(}{)}{%
  #1\;\delimsize|\delimsize|\;#2%
}
\newcommand{\kld}[2]{\ensuremath{D_{KL}\infdivx{#1}{#2}}\xspace}

\usepackage[linesnumbered,ruled,vlined]{algorithm2e}
\SetKwInput{KwInput}{Input}
\SetKwInput{KwOutput}{Output}

\newcommand{\poultrybarn}{\textsc{Poultry barn}\xspace}
\newcommand{\solarfarm}{\textsc{Solar farm}\xspace}

\begin{document}

\title{Temporal Cluster Matching for Change Detection of Structures from Satellite Imagery}

\author{Caleb Robinson}
\email{caleb.robinson@microsoft.com}
\affiliation{%
  \institution{Microsoft AI for Good Research Lab}
}

\author{Anthony Ortiz}
\email{anthony.ortiz@microsoft.com}
\affiliation{%
  \institution{Microsoft AI for Good Research Lab}
}

\author{Juan M. Lavista Ferres}
\email{jlavista@microsoft.com}
\affiliation{%
  \institution{Microsoft AI for Good Research Lab}
}

\author{Brandon Anderson}
\email{branders@law.stanford.edu}
\affiliation{%
  \institution{Stanford RegLab}
}

\author{Daniel E. Ho}
\email{dho@law.stanford.edu}
\affiliation{%
  \institution{Stanford RegLab}
}

\renewcommand{\shortauthors}{Robinson, et al.}

\begin{abstract}
Longitudinal studies are vital to understanding dynamic changes of the planet, but labels (e.g., buildings, facilities, roads) are often available only for a single point in time. We propose a general model, Temporal Cluster Matching (TCM), for detecting \emph{building changes} in time series of remotely sensed imagery when footprint labels are observed only once. 
The intuition behind the model is that the relationship between spectral values inside and outside of building's footprint will change when a building is constructed (or demolished). For instance, in rural settings, the pre-construction area may look similar to the surrounding environment until the building is constructed. Similarly, in urban settings, the pre-construction areas will look different from the surrounding environment until construction.
We further propose a heuristic method for selecting the parameters of our model which allows it to be applied in novel settings without requiring data labeling efforts (to fit the parameters).
We apply our model over a dataset of poultry barns from 2016/2017 high-resolution aerial imagery in the Delmarva Peninsula and a dataset of solar farms from a 2020 mosaic of Sentinel 2 imagery in India.
Our results show that our model performs as well when fit using the proposed heuristic as it does when fit with labeled data, and further, that supervised versions of our model perform the best among all the baselines we test against.
Finally, we show that our proposed approach can act as an effective data augmentation strategy -- it enables researchers to augment existing structure footprint labels along the time dimension and thus use imagery from multiple points in time to train deep learning models. We show that this improves the spatial generalization of such models when evaluated on the same change detection task.
\end{abstract}



\maketitle

\section{Introduction}

Catalogs of high-resolution remotely sensed imagery have become increasingly available to the scientific community. The availability of such imagery has revolutionized scientific fields and society at large. For example, 1m resolution aerial imagery from the US Department of Agriculture (NAIP imagery) has been released on a 2-year rolling basis over the entire US for over a decade and the commercial satellite imagery provider, Planet, recently started to release 5m satellite imagery covering the whole tropical forest region of the world on a monthly basis. One estimate is that the opening of Landsat imagery in 2008 led to the creation of \$3.45B in economic value in 2017 alone~\cite{Straub2019}. The accumulation of such data facilitates an entirely new branch of longitudinal studies -- analyzing the Earth and how it has changed over time. 

As the climate, technology, and human population change on an ever more rapid timescale, such longitudinal studies become particularly vital to understanding the past, present, and future of the environment.  Despite the usefulness of time series data, such research faces two important practical challenges. First, the large labeled datasets that have fueled advances in computer vision are much more limited in the satellite imagery context  \cite{DBLP:journals/corr/abs-1807-01232, roscher2020semcity, demir2018deepglobe}. Second, efforts in creating labeled data from remotely sensed imagery are typically focused on a single point in time \cite{DBLP:journals/corr/abs-1807-01232, weir2019spacenet, roscher2020semcity}. Project requirements may only call for a single layer of labels, or budget constraints may limit the number of labels that can be generated. This has the effect of creating labeled datasets that are ``frozen'' in time.
Expanding such ``frozen'' datasets to multiple points in time in independent follow-up work can be resource-intensive and difficult, as the same image-preprocessing and labeling methodology steps used in the original work need to be precisely reproduced in order to generate comparable data.

Going beyond ``frozen'' datasets would enable a wide range of \emph{temporal} inferences from satellite imagery, with significant social, economic, and policy implications. Previous studies include the detection of urban expansion \cite{wang2012china}, zoning violations \cite{purdy2010using}, habitat modification \cite{evans2020automated}, compliance with agricultural subsidies \cite{moltzau}, construction on wetlands  \cite{handan2020deep}, and damage assessments from natural disasters \cite{gupta2020rescuenet, gupta2019cnn, matsuoka2004use}.  

Algorithmic approaches for expanding ``frozen'' datasets can thus be useful in facilitating ecological and policy-based analysis.  In this work we propose a simple model, Temporal Cluster Matching (TCM), for determining \textit{when} structures were previously constructed given a labeled dataset of structure footprints generated from imagery captured at a particular point in time. This model, importantly, does not rely on the differences in spectral values between layers of remotely sensed imagery as there can be considerable variance in these values depending on imaging conditions, the type of sensor used, etc. Instead, it compares a representation of the spectral values inside a building footprint to a representation of the spectral values in the surrounding area for each point in the time series. Whenever the distribution of spectral values within the footprint becomes dissimilar to that of its surroundings then the footprint is likely to have been developed. We further propose a method for fitting the parameters of this model which does not rely on additional labeled footprint data over time and show that this ``semi-supervised TCM'' performs comparably to supervised methods.

Specifically, we demonstrate the performance of this algorithm in two distinct settings:
\begin{enumerate}
    \item Poultry barns from concentrated animal feeding operations (CAFOs) in the United States, using high-resolution aerial imagery from the National Agricultural Imagery Program (NAIP), and
    \item Solar farm footprints in the Indian state of Karnataka, using Sentinel 2 annual mosaics.
\end{enumerate}

Both settings are of significant environmental importance and are ripe for longitudinal study.  First, CAFOs can have profound effects on water quality and human health in their proximity \cite{doi:10.2105/AJPH.94.10.1703}. Nitrates and other potentially harmful chemicals can, for example, make their way into the groundwater, spreading to adjacent wells and bodies of water over timescales that range into decades.  Usage of antibiotics for growth promotion can lead to resistant bacterial infections in nearby populations \cite{cafo-antibiotics}. Effective regulation in either scenario requires differentiation of these contaminant sources, which, in turn, depends on accurate historical labels and spatio-temporal modeling.

Second, understanding the growth of solar systems is increasingly important in the transition toward clean energy.
India is an important example of this, as it has set ambitious goals of generating 450 GW of renewable energy by 2030 with 175 GW deployment by 2022 \cite{frangoul}. 
Achieving this goal will require an expansion of solar farm installations throughout the country and policy makers will be able to determine better the effects of country-wide efforts with solar farm change data that can be updated year-over-year in a consistent manner. Understanding such solar expansion may also enable more targeted investments for solar potential \cite{Moynihan}.

To summarize, our contributions include:
\begin{itemize}
    \item A lightweight model, Temporal Cluster Matching, for detecting when structures are developed in a time-series of remotely sensed imagery, as well as a heuristic method for fitting the parameters of the model. Combined, this results in a proposed approach that only relies on labeled building footprints for a single point in time.
    \item A series of baseline methods, both supervised and semi-supervised, to evaluate our proposed approach against. 
    \item Experiments comparing our model to the baseline models in two datasets: poultry barn footprints with aerial imagery, and solar farm footprints with satellite imagery.
    \item A code release that allows users to run the model in novel settings, as well as scripts for reproducing our experiments: \url{https://github.com/microsoft/temporal-cluster-matching}
\end{itemize}

\begin{figure*}[t!]
    \begin{center}
    \includegraphics[width=0.9\linewidth]{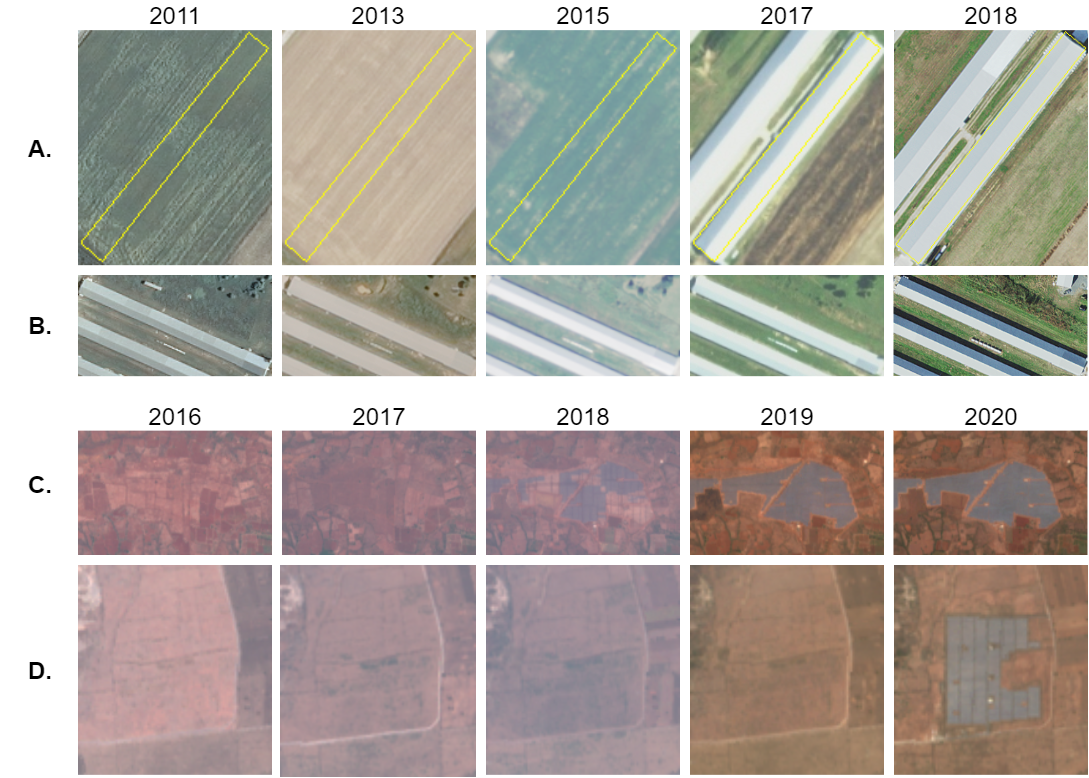}
    \end{center}
    \caption{(\textbf{A} and \textbf{B}) Examples of two poultry barn footprints over 5 years of NAIP imagery. We observe inter-year variability of NAIP imagery and the change in the relation of color/texture between the footprint and neighborhood when a footprint is ``developed''. (\textbf{C} and \textbf{D}) Examples of two solar farm footprints over 5 years of Sentinel 2 imagery. Note, in \textbf{A} we outline the building footprint location in yellow through the entire series of imagery, but omit this outline in remaining rows.}
    \label{fig:data-examples}
\end{figure*}

\section{Related Work}
Our work pertains to several different literatures.
First, much work has focused on methods for detection of building footprints. For instance, \cite{zhao2018building} uses Mask R-CNN to train a model to detect buildings while \cite{yang} uses a semantic segmentation model (U-Net \cite{ronneberger2015u}) to segment buildings in imagery. While deep learning approaches have made rapid advances, they have largely been focused on static inferences. Moving to a different domain (spatially or temporally) can prove challenging due to shifts in the input distribution (what the input images look like), co-registration errors, and shifts in the target (what buildings looks like)~\cite{tuia2016domain}. Zhang et al., for instance, note these as some of the challenges faced when detecting structures at a global scale~\cite{zhang2017building}. 

Second, other research has focused on detecting changes in satellite imagery. Historically, the remote sensing literature has started from pixel-wise change-point detection -- detecting when change happens in a time series of repeated observations of the same location in space. Much work has focused on how to model the characteristics of these time series such as seasonality, changes in illumination, atmospheric conditions, etc~\cite{aminikhanghahi2017survey}. A popular method for performing this task, BFAST, models a time series of remotely sensed observations with trend, seasonal, and remainder components, then uses an unsupervised iterative method to detect change points based on the model~\cite{verbesselt2010detecting}. This method relies on observing a relatively long time series, e.g. that shows seasonal components, and thus is not applicable to time series with few data points. Change-point detection can also be performed on other units of analysis. \cite{tewkesbury2015critical} review the literature and organize methods based on their unit of analysis, e.g. pixel-based, object-based, kernel-based, and based on their method for comparing scenes, e.g. based on differencing, transformation, or modeling. Within this organization, our work is the most similar to those that operate on image-object overlays~\cite{tewkesbury2011mapping,listner2011recent,huang2020automatic} whereby a single segmentation is applied to all imagery in a time series and change at an object level is computed. We use similar ideas in designing our baseline approaches (Section \ref{subsec:baselines}).

The computer vision and machine learning literature also addresses similar problems. Several methods use a fully supervised approach to detect changes in known building locations: \cite{jung2004detecting} use a supervised approach with decision trees to classify whether a building change occurred from pairs of imagery and \cite{malpica2013change} use support vector machines to provide estimates for which buildings have changed. Other methods perform a superpixel segmentation step to create objects in pairs of imagery, then model change over these objects with a Markov random field~\cite{marcos2016geospatial}. Most recently, advances in deep learning have driven end-to-end pipelines in building change detection. \cite{chen2019changenet}, for instance, uses a Generative Adversarial Network (GAN) to overcome the limitations of pixel-level inferences. These methods all rely on having existing labeled data on either when changes have occurred, or on unchanged areas in consecutive pairs of imagery. 

Finally, numerous research teams have provided benchmark datasets for evaluating models at a single point in time  \cite{DBLP:journals/corr/abs-1807-01232, weir2019spacenet, roscher2020semcity}.  Few datasets provide longitudinal information about the same location over time, so a typical research approach has been to train a model on labeled imagery from one period. \cite{handan2019deep}, for instance, assess the growth of intensive livestock farms in North Carolina, but do so using a model trained on images of such facilities for a single period of time. A notable exception to this is the recent SpaceNet 7 dataset/challenge~\cite{van2021multi}. This dataset includes 24 multi-spectral images at a 4m/px spatial resolution as well as building footprint labels over time for over 100 unique locations around the world. It is particularly challenging for object based change detection approaches as the median building size is 12.1 pixels~\cite{van2021multi} and the imagery is not perfectly co-registered. Pixel based segmentation models followed by in-depth post-processing methods achieved the top performance in the competition~\cite{van2021spacenet}, however it is not yet clear how to adapt such methodology to general change detection tasks.

Our approach contributes to this body of work by providing a semi-supervised approach for detecting changes in structures that easily enables researchers to expand a dataset beyond a single time period, hence enabling domain adaptation by efficient sampling of images across time. The approach we propose can be seen as a lightweight, data-driven method to expand ``frozen'' imagery longitudinally, enabling researchers to address a rich set of  dynamic questions.

\begin{figure*}[th]
    \centering
    \includegraphics[width=0.85\linewidth]{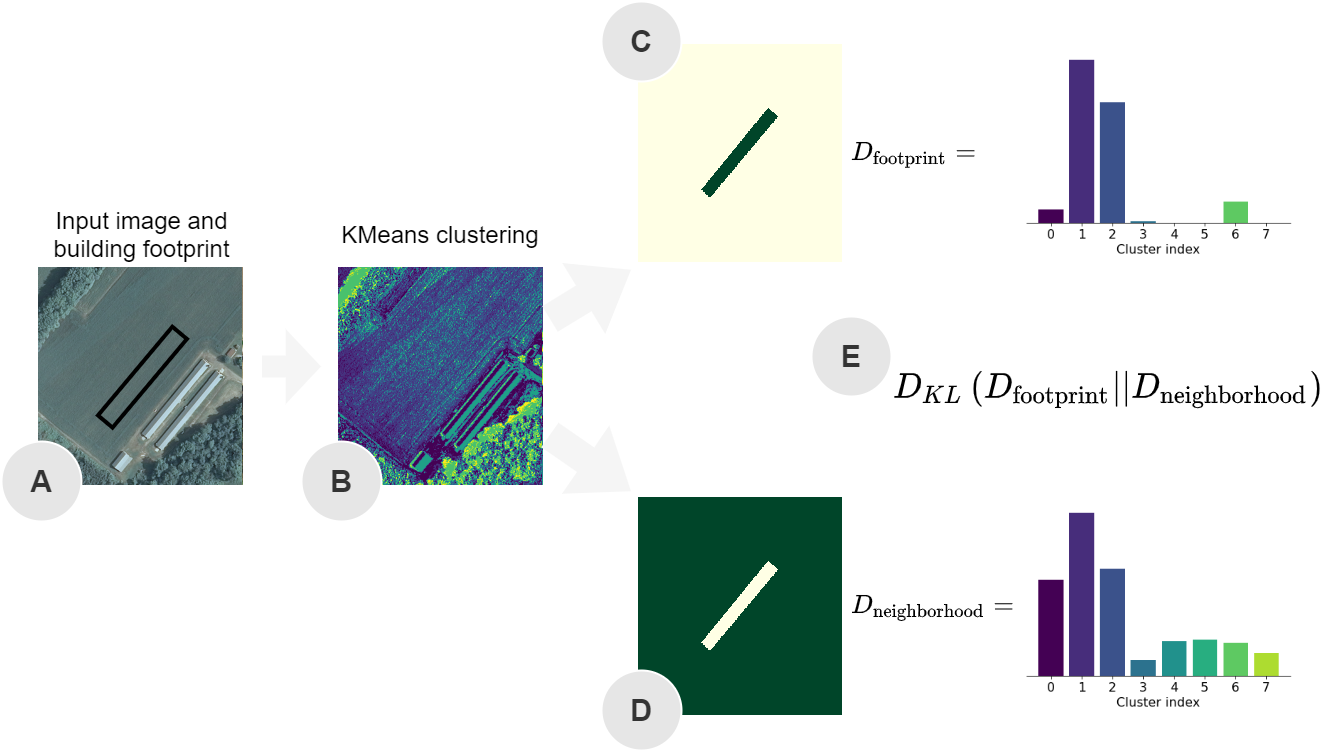}
    \caption{Schematic description of Temporal Cluster Matching: (\textbf{A}) the input imagery and query building footprint; (\textbf{B}) $k$-means clustering of the input imagery (\textbf{C, D}) discrete distributions of clusters created from the pixels within the \textbf{footprint} polygon and from the pixels outside of the footprint (i.e. the \textbf{neighborhood});  (\textbf{E}) the KL-divergence between the two distributions indicates the similarity of the footprint to its neighborhood.}
    \label{fig:schematic}
\end{figure*}

\section{Methods}

\subsection{Problem statement}
Formally, we would like to find when a structure was developed (or, more generally, changed) given a time series of remotely sensed imagery of it and the surrounding area, $\left[ \boldx^{1}, \ldots, \boldx^{t} \right]$ and its \textit{footprint}, $P$, at time $t$. We represent this footprint as a mask, $\boldy^t$. Here, the $i^{\text{th}}$ image in the time series $\boldx^i \in \mathbb{Z}^{w \times h \times c}$ is a georeferenced image with width, $w$, height, $h$, and number of spectral bands, $c$. Similarly, $\boldy^t \in \{0, 1\}^{w \times h}$ is a georeferenced binary mask with the same dimensions that contains a $1$ in every spatial location that the given structure covers at time $t$ and a $0$ elsewhere. We want to estimate the point in time that the structure was created, i.e. find $l \in \left[1, t\right]$, where $\boldx^{l}$ contains the structure, for the smallest such $l$. Note that we assume the structure to exist at $t$.

\subsection{Temporal Cluster Matching}

Our proposed model, Temporal Cluster Matching, relies on the assumption that when a structure is built its footprint will have a different set of colors and textures than its immediate surroundings compared to when the structure was not built. For example, an undeveloped piece of land in a rural setting will likely contain some sort of vegetation, and that vegetation will probably look similar (in color/texture) to some of its surroundings. When a structure is built on this land, then it will likely look dissimilar to its surroundings (unless e.g. its entire surroundings are also developed at the same time). The same intuition holds in urban environments -- an undeveloped piece of land will look dissimilar to its surroundings, however, when it is developed, it will look similar.\\
\\
We assume that we are given a \textit{footprint}, $P$, that outlines a structure that has been labeled as developed at time $t$. Now, we formally define the \textit{neighborhood} of this \textit{footprint}. This \textit{neighborhood} should be larger than the extent of the footprint in order to observe a representative set of colors/textures, so we let $r$ be a radius that serves as a buffer to the building footprint polygon. We then create $\boldy^t$ by rasterizing the polygon within this buffered extent and create $\left[ \boldx^{0}, \ldots, \boldx^{t} \right]$ by cropping the same buffered extent from each layer of remotely sensed imagery.\\
\\
Next, we define a method for comparing the set of colors/textures within the \textit{footprint} to those in the surrounding \textit{neighborhood}. Given a single layer of remotely sensed image from the time series, $\boldx$, we run $k$-means to partition the pixels into $k$ clusters. Each pixel can be represented by a set of features that encodes color and texture at its location, for example: the spectral values at the pixel's location, a texture descriptor (such as a local binary pattern) at the location, the spectral values in a window around the location, or some combination of the previous representations. Regardless, the cluster model will assign a cluster index to each pixel in $\boldx$ which we call $\mathbf{C}$. We then represent an area by the discrete distribution of cluster indices observed in that area. Specifically, we let $D_\text{footprint}$ be the distribution of cluster indices from $\mathbf{C}[\boldy^t = 1]$ and $D_\text{neighborhood}$ be the distribution of cluster indices from $\mathbf{C}[\boldy^t = 0]$\footnote{We use the notation $\mathbf{C}[\boldy^t = 1]$ to mean all the cluster indices of pixels where $\boldy^t = 1$. We build the discrete distribution by counting the number of pixels assigned to each cluster and normalizing the vector of counts by its sum.}. Now, we can compare the set of colors/textures within a footprint to those in its surrounding \textit{neighborhood} by calculating the KL-divergence between the two distributions of cluster indices, $d = \kld{D_\text{footprint}}{D_\text{neighborhood}}$. Larger KL-divergence values mean that the color/texture of a footprint is dissimilar to that of its surrounding neighborhood and that it is likely to be developed. We perform this comparison method for each image in the time series to create a list of KL-divergence values $\left[d_1, \ldots, d_t\right]$\\
\\
Finally, we let $\theta$ be a threshold value to determine the smallest KL-divergence value that we will consider to indicate a ``developed'' footprint. More specifically, our model will estimate $l$ as the time that a footprint is first developed for the first $l$ where $d_l > \theta$. This parameter can be found by experimentation using labeled data, or with the heuristic method we describe in Section \ref{subsec:heuristic}. See Algorithm \ref{pseudocode} and Figure \ref{fig:schematic} for an overview of this proposed approach.

\begin{algorithm}[thb]
\SetAlgoLined
\KwInput{Time series of remotely sensed imagery, $P$, $k$, $r$, and $\theta$}
\KwOutput{$l$, the \textit{first} point in time that the footprint described by $P$ was developed}
$\left[ \boldx^{1}, \boldx^{1}, \ldots, \boldx^{t} \right] \gets$ crop the imagery according to the buffered extent of $P$ by a radius $r$ \\
$\boldy^t \gets$ rasterize $P$ in the same buffered extent\\
\For{$l\gets1$ \KwTo $t$}{
    $\mathbf{C} \gets$ cluster indices from a $k$-means clustering of $\boldx^{l}$ into $k$ clusters
    
    $D_\text{footprint} \gets$ distribution of cluster indices  $\mathbf{C}[\boldy^t = 1]$
    
    $D_\text{neighborhood} \gets$ distribution of cluster indices  $\mathbf{C}[\boldy^t = 0]$
    
    $d \gets \kld{D_\text{footprint}}{D_\text{neighborhood}}$ \\
    \If{$d > \theta$}
    {
        return $l$
    }
}
return $t$
\caption{Temporal Cluster Matching}
\label{pseudocode}
\end{algorithm}

Section \ref{subsec:baselines} explores more complex decision models than the single threshold described above, but we note that these require labeled data to fit.

\subsection{A heuristic for semi-supervised Temporal Cluster Matching}
\label{subsec:heuristic}
In application scenarios we would like to use our model, given a dataset of (a) known structure footprints at time $t$ and (b) a time series of remotely sensed imagery over a certain study area, to find when each structure was constructed. Here we propose a method for determining reasonable parameter values for the number of clusters, $k$, buffer radius, $r$, and decision threshold, $\theta$, \textit{without assuming that we have prior labeled data on construction dates}.

\begin{figure}[ht]
    \centering
    \includegraphics[width=0.85\linewidth]{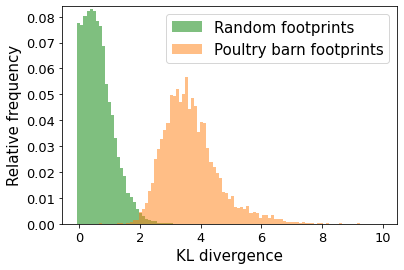}
    \caption{We show the distribution of KL divergence values generated by our proposed approach for (1) poultry barn footprints in aerial imagery where there is guaranteed to be a structure and (2) randomly generated footprints over similar aerial imagery. We observe that (1) is unimodal with a large mean value, as the color distributions of footprints that contain buildings are dissimilar to the color distribution of their surroundings and (2) is unimodal with a small mean value, as random patches are highly likely to have a color distribution that is similar to their neighborhood. Our proposed heuristic method looks to find hyperparameters for the model that minimize the overlap in these two distributions such that a simple threshold can identify changes in building footprints.}
    \label{fig:known_vs_random}
\end{figure}

This heuristic compares the distribution of KL-divergence values calculated by our algorithm for given hyperparameters, $k$ and $r$, over all footprints at time $t$ (when we assume that structures exist) to the distribution of KL-divergence values over a set of \textit{randomly generated polygons} over the study area. The intuition is that the relationship between random polygons and their \textit{neighborhoods} is similar to the relationship between undeveloped structure footprints and their \textit{neighborhoods}. In other words, this distribution of KL divergence values between color distributions from random polygons and their surroundings will represent what we would expect to observe by chance -- i.e. \textit{not} the relationship between the colors in a building footprint and its surroundings. We want to find parameter settings for our algorithm that minimize the overlap between these two distributions because it will make it easier to identify change (see Figure \ref{fig:known_vs_random} for an illustration of this for poultry barn footprints). Formally, we let $p$ be the distribution of KL-divergence values over footprints at $t$ and $q$ be the distribution of KL-divergence values over random polygons sampled from the study area (over all points in time). These are discrete distributions (e.g after binning KL divergence values) and we can measure the overlap with the Bhattacharyya coefficient, $BC(p, q) = \sum_{x \in X} \sqrt{p(x) q(x)}$. Choosing $k$ and $r$ thus becomes a search $\min_{k, r} BC(p, q)$.

Finally, after choosing $k$ and $r$, we can simply choose $\theta$ as a value representing the 98th percentile (or similar) of the resulting distribution of random polygons, $q$. Practically, this value simply needs to separate $p$ and $q$ and visualization of these two distributions should suggest appropriate values.

We test this heuristic in Section \ref{subsec:heuristic-experiments} by comparing change detection performance from fitting our proposed model with this heuristic versus with labeled data.

\subsection{Baseline approaches}
\label{subsec:baselines}

Here we propose a series of baselines and variants of our model to compare against. We refer to our proposed model / heuristic for fitting the model as ``Semi-supervised TCM'' as it only depends on labeled building footprints from a single point in time. This is specifically in contrast to variant approaches like ``Supervised TCM'' (see below) that use labeled building footprints over time to fit the model parameters.


\begin{description}
\item[Supervised TCM] Here, we fit the parameter $\theta$ in our proposed model using labeled data instead of our proposed heuristic. We can do this by searching over values of $\theta$ and measuring performance on the labeled data. In this case, $k$ and $r$ are model hyperparameters that can be searched over using validation data.
\item[Supervised TCM with LR] We use the series of KL-divergence values computed by TCM as a feature representation in a logistic regression (LR) model that directly predicts which point in time a structure is first observed. This is a supervised method as it requires a sample of labeled footprint data over time to fit.
\item[Average-color with threshold] This baseline uses the same structure as TCM with two changes: instead of clustering colors we compute average colors representations (over space for each spectral band) and instead of computing KL-divergence between distributions of cluster indices we compute the Euclidean distance between the average colors representations. Specifically, we compute the average color in a footprint and the average color of its neighborhood, then take the Euclidean distance between them and treat this distance in the same way we have previously treated the KL-divergence values. This has the effect of removing $k$ as a hyperparameter, however the rest of the algorithm stays the same. Similar to the KL with threshold method we fit $\theta$ using labeled data.
\item[Average-color with LR] This method is identical to Supervised TCM with LR, but using the technique from Average-color with threshold to compute Euclidean distances between average color representations.
\item[Color-over-time] In this baseline we compute features from a time series of imagery by averaging the colors (over space for each spectral band) in the given footprint at each point in time, then taking the Euclidean distance between these average representations in subsequent pairs of imagery. For example, a time series of 5 images would result in an overall feature representation of 4 distances: the distance between the average colors at time 1 and average colors at time 2, the distance between the average colors at time 2 and the average colors at time 3, etc. We use this overall representation in a logistic regression model that predicts which point in time the structure is first observed.
\item[CNN-over-time] In this baseline we use the given structure footprints and satellite imagery at time $t$ to train a U-Net based semantic segmentation model to predict whether or not each pixel in an image contains a developed structure. We then use this trained model to score the imagery from each point and time and determine the first layer in which a building is constructed. For simplicity, if the network predicts that over 50\% of a footprint is constructed, then we count it as constructed.
\item[Mode predictions] This baseline is simply predicting the most frequent time point that we first observe constructed buildings based on the labels in the dataset of interest. For example, if `2011' was the most frequent year that we observed buildings to be constructed in a dataset, then this approach would predict every building was constructed in 2011 regardless of input. This serves as a lower bound on the performance of the supervised methods.
\end{description}

\section{Datasets}

\subsection{Poultry barn dataset}

We use the Soroka and Duren dataset of 6,013 labeled poultry barn polygons, \poultrybarn, created from aerial imagery from the National Agriculture Imagery Program (NAIP) from 2016/2017 over the Delmarva Peninsula (containing portions of Virginia, Maryland, and Delaware)~\cite{sorokaDataset}. NAIP imagery is 4 channel (red, green, blue, NIR) high-resolution ($\leq 1\text{m/px}$) aerial imagery and is collected independently by each state in the US at least once every three years on a rolling basis. Because of this, the availability and quality of the imagery varies between states. For instance, the NAIP imagery from 2011 in Delaware and Maryland are collected on different days of the year, at different times of day, etc. See Figure \ref{fig:data-examples} for example images of the NAIP imagery over time overlayed with the barn footprints. Additionally, we have manually labeled the earliest year (out of the years shown in Table \ref{tab:poultry-data}) that a poultry barn can be seen for a random subset of 1,000 of the poultry barn footprints.

\begin{table}[ht]
\begin{center}
\begin{tabular}{ll}
\toprule
\multicolumn{1}{c}{\textbf{State}} & \multicolumn{1}{c}{\textbf{Years of NAIP data}} \\ \midrule
Delaware                           & 2011, 2013, 2015, 2017, 2018                        \\
Maryland                           & 2011, 2013, 2015, 2017, 2018                        \\
Virginia                           & 2011, 2012, 2014, 2016, 2018                        \\ \bottomrule
\end{tabular}
\end{center}
\caption{NAIP data availability over states covering the Delmarva Peninsula.}
\label{tab:poultry-data}
\end{table} 

\subsection{Solar farm dataset}

We also use a solar farm dataset, \solarfarm, containing polygons delineating solar installations in the Indian state of Karnataka for the year 2020. The dataset includes 935 individual polygons covering a total area or 25.7 km\textsuperscript{2}. The polygons were created by manually filtering the results of a model run on an annual median composite of Sentinel 2 multispectral surface reflectance imagery. We collect additional median composites of Sentinel 2 imagery for 2016 through 2019\footnote{The data from 2016, 2017 and 2018 are composites of the Sentinel 2 top of atmosphere products (Level 1C), while the 2019 and 2020 data are additionally corrected for surface reflectance (Level 2A). All data was processed with Google Earth Engine using the \texttt{COPERNICUS/S2} and \texttt{COPERNICUS/S2\_SR} collections respectively.} to use for change detection. See Figure \ref{fig:data-examples} for examples of the imagery overlayed with the solar farm footprints. For each of the 935 footprints we have manually labeled the earliest year (between 2016 and 2020) that a solar farm can be seen in the imagery. 

\begin{table*}[ht]
\begin{center}
\begin{tabular}{@{}llcrr@{}}
\toprule
                              & \multicolumn{1}{c}{\textbf{Method}} & {\textbf{Semi-Supervised}} & \multicolumn{1}{c}{\textbf{ACC}} & \multicolumn{1}{c}{\textbf{MAE}} \\ \midrule
\multirow{1}{*}{\poultrybarn} & Semi-supervised TCM                 & \checkmark      & 0.94                             & 0.15                             \\
                              & Supervised TCM                      &                 & 0.93 +/- 0.01                    & 0.17 +/- 0.04                    \\
                              & Supervised TCM with LR              &                 & 0.96 +/- 0.01                    & 0.12 +/- 0.03                    \\
                              \cline{2-5}
                              & CNN over time                       & \checkmark      & 0.37                             & 1.36                             \\ 
                              & Average-color with LR               &                 & 0.95 +/- 0.01                    & 0.15 +/- 0.05                    \\
                              & Average-color with threshold        &                 & 0.91 +/- 0.02                    & 0.24 +/- 0.06                    \\
                              & Color-over-time                     &                 & 0.90 +/- 0.02                    & 0.41 +/- 0.08                    \\
                              & Mode predictions                    &                 & 0.84                             & 0.80                             \\ \midrule
\multirow{1}{*}{\solarfarm}   & Semi-supervised TCM                 & \checkmark      & 0.71                             & 0.49                             \\
                              & Supervised TCM                      &                 & 0.70 +/- 0.04                    & 0.51 +/- 0.08                    \\
                              & Supervised TCM with LR              &                 & 0.78 +/- 0.03                    & 0.29 +/- 0.05                    \\
                              \cline{2-5}
                              & CNN over time                       & \checkmark      & 0.64                             & 0.68                             \\
                              & Average-color with LR               &                 & 0.65 +/- 0.03                    & 0.49 +/- 0.05                    \\
                              & Average-color with threshold        &                 & 0.50 +/- 0.04                    & 0.93 +/- 0.08                    \\
                              & Color-over-time                     &                 & 0.79 +/- 0.01                    & 0.29 +/- 0.02                    \\
                              & Mode predictions                    &                 & 0.42                             & 0.81                             \\ \bottomrule
\end{tabular}%
\end{center}
\caption{Comparison of our proposed semi-supervised model (``Semi-supervised TCM'') to other baseline methods for detecting change in structures over time series of imagery. Note that the semi-supervised methods only have access to building footprint labels at time $t$, while the other methods are ``supervised'' and additionally have access to labels on when buildings were constructed over time. We observe that our semi-supervised approach achieves identical performance to a supervised variant where the model parameters are learned. We further observe the proposed approach outperforms supervised baseline methods for detecting change. Reported values are shown as averages (+/-) a standard deviation over 50 random train/test splits where appropriate. Single values are reported for the semi-supervised methods as they are evaluated on the entire labeled data set.}
\label{tab:results}
\end{table*}

\section{Experiments and results}

We experiment with different configurations of our algorithm on the \poultrybarn and \solarfarm datasets. In all experiments we measure the accuracy (ACC) -- the percentage of labeled footprints for which we correctly identify the first ``developed'' year and mean absolute error (MAE) -- the average of absolute differences between the predicted year and labeled year.

\subsection{Semi-supervised TCM}
\label{subsec:heuristic-experiments}

We first test how parameters chosen with the proposed heuristic correlate with performance of the model. The benefit of the heuristic method is that it does not require labeled temporal data to fit the model, but we need to show that the parameters it selects actually result in good performance. Here, we search over buffer sizes in $\{100, 200, 400\}$ meters and $\{0.016, 0.024\}$ degrees for the \poultrybarn and \solarfarm datasets, respectively, and number of clusters in $\{16, 32, 64\}$ for both datasets. For each configuration combination we create $p$ and $q$ as described in Section \ref{subsec:heuristic}, compute the Bhattacharyya coefficient, estimate $\theta$, then evaluate the predicted change years on the labeled data. We find that the Bhattacharyya coefficient is correlated with the result; there is -0.77 rank order correlation between the coefficient and accuracy (p=0.01) in \poultrybarn and a -0.94 rank order correlation (p=0.004) in \solarfarm. In both datasets, the smallest Bhattacharyya coefficient was paired with the best performing algorithm configuration.

Second, we compare the performance of the model with heuristic estimated parameters to that with learned parameters. To learn the parameters for our proposed model, ``Supervised TCM'',  we randomly partition the labeled time series data into 80/20 train/test splits. We find the values of $k$, $r$, and $\theta$ (with a grid search over the same space for $k$ and $r$ as mentioned above) using the training split, then evaluate this model on the test split. We repeat this process for 50 random partitions and report the average and standard deviation metrics for the best combination Table \ref{tab:results}. We observe that the heuristic method produces results that are equivalent to those of the learned model. In \poultrybarn our proposed method achieves a 94\% accuracy with a mean absolute error of 0.15 years which suggests it will be effective for driving longitudinal studies of the growth of poultry CAFOs.

Finally, we observe that our method significantly outperforms the other semi-supervised baseline, CNN over time. In both \poultrybarn and \solarfarm we observe considerable covariate shift. For example, in the \poultrybarn dataset there is a large shift in input distribution over time due to the fact that the aerial imagery is collected at different days of the year, at different times of day, etc. The deep learning model is trained solely on imagery from the last point in each time series where we can confirm that there exists buildings in each footprint, however is unable to reliably generalize over time. We did not experiment with domain adaptation techniques to attempt to fix this, however we explore the use of our proposed method in this capacity in Section \ref{sec:dl}. We note that our proposed model is unaffected by shifts in the input distributions year-over-year as it never compares imagery from different years.

\subsection{Supervised models}
In the previous section we showed that we can estimate the parameters of our model without labeled time series data. Not requiring additional labeling is a major benefit of the TCM approach. In this section we explore the performance of our proposed approach against \textit{supervised} baseline approaches, using labels generated going back in time. We find that logistic regression models are effective at predicting the building construction date from the series of KL divergence values produced by our proposed approach (KL with LR). In both datasets this method is overall the top performing method with 96\% accuracy in the \poultrybarn dataset and 78\% accuracy in the \solarfarm dataset. In the \solarfarm dataset the Color-over-time baseline has tied for top performance (within a standard deviation), but the same features are not as effective in the \poultrybarn dataset where the color shifts are more dramatic year-over-year. Even so, the performance of the color-over-time baseline was much better than we originally hypothesized and should be compared to in future work regardless of perceived covariate shifts.

We also observe that our proposed method (that computes clustered representations) dominates the family of average-color baselines. This, along with the fact that we observe that more clusters in the $k$-means model usually results in better performance, suggests that the clustered representation is an important component of our approach. We hypothesize that more rich feature representations will prove even more effective as both the colors and textures of a footprint will change when it becomes developed. This is a trivial addition to the existing model and we hope to test it in future studies. 

Finally, we observe that our heuristic method performs very well overall. In both datasets there are only two supervised methods that achieve stronger results than the semi-supervised proposed approach.

\section{Temporal Cluster Matching as a data augmentation strategy}
\label{sec:dl}

In Table \ref{tab:results} we show that training semantic segmentation models on structure footprint masks at a time $t$ does \textit{not} result in a model that can generalize well over time (and thus cannot detect change). Previously (in Section \ref{subsec:heuristic-experiments}) we hypothesized that this is due to the covariate shift in the time series imagery in the two datasets that we test on -- the \poultrybarn dataset uses NAIP aerial imagery that is collected at different times of day and different days of the year on a rolling three year basis and the \solarfarm dataset uses Sentinel 2 mosaics created from TOA corrected imagery in 2016 through 2018 and surface reflectance corrected imagery in 2019 and 2020. Thus, a model trained with data from a single period in both of these cases is unlikely to perform well in other layers.

Here we test this hypothesis by using our proposed method to augment the data used to train the CNN over time method for the \poultrybarn dataset. Specifically, we run Semi-supervised TCM over the \poultrybarn dataset to create predictions as to when each footprint was constructed. We then use these estimates to create an expanded training set that contains pairs of imagery over all time points with footprint masks that are predicted to have a building. For example, if our model believes that there was a building in a given footprint at 2011 in the NAIP imagery, then we can train the segmentation model with (NAIP 2011, footprint), (NAIP 2013, footprint), etc. We find that this increases the performance of the model on the change detection task in all cases that we tested. For example, we apply this augmentation step to the same model configuration used in the results from Table \ref{tab:results} and achieve a 56\% accuracy and 0.97 MAE (a 19\% improvement in ACC and 0.39 improvement in MAE). These results are not competitive with the other methods we test in the change detection task.  This likely stems from the fact that the segmentation model, in contrast to the other methods, is not specialized to  change detection. On the other hand, the segmentation model can be run over new imagery to find novel instances of poultry barns (e.g., barns destroyed prior to the date of original data labeling), and is thus necessary to improve the performance of such models for more general applications.

While a more rigorous evaluation of how to improve the deep learning segmentation baseline is outside the scope of this paper, we hypothesize that more data augmentation strategies (e.g. RandAugment~\cite{cubuk2020randaugment} and AugMix~\cite{hendrycks2019augmix}), unsupervised domain adaptation methods~\cite{sun2019unsupervised}, and a hyperparmeter search over dimensions such as class balancing strategies, temporal balancing strategies, learning rates, architecture,  etc. would all improve performance. These types of experimentation will be critical for any future work that attempts to create general purpose models for detecting building construction at scale.  That said, one of the main benefits of our proposed TCM is that it provides a lightweight approach to detect construction.

\section{Conclusion and future work}
We have proposed  Temporal Cluster Matching (TCM) for detecting change in building footprints from time series of remotely sensed imagery. This model is based on the intuition that the relationship between the distribution of colors inside and outside of a building footprint will change upon construction. We further propose a heuristic based method for fitting the parameters of our model and show that this approach effectively detects poultry barn and solar farm construction. TCM does not depend on having labels over time, yet it can outperform similar models that have such labels available. Further, we show that the feature representation from TCM -- a sequence of KL-divergence values between the distribution of color clusters inside and outside of a building footprint -- can be used in supervised models to improve change detection performance. Finally, we show how TCM can be used as a data augmentation technique for training deep learning models to \textit{detect} building footprints from remotely sensed imagery.

This work motivates several future directions.
First, the per-pixel representation of TCM will affect detectable changes. We used simple color representations, but more elaborate representations could be promising (e.g. texture descriptors or higher dimensional image embeddings).
Second, future work should explore other applications of TCM. Here we experimented with imagery where the size of the footprints were relatively large compared to the spatial resolution of the imagery. However, our model may not perform as well when the footprint is relatively smaller. For example, we briefly experimented with detecting changes using general building footprints in the US and NAIP imagery and found that the relationship between the color distributions of small residential buildings and their surroundings was very noisy, although we did not attempt to investigate further. The top performing methods from the recent SpaceNet7 challenge run their building detection algorithms on upsampled imagery and a similar strategy may be useful with TCM.
Finally, we hypothesize that TCM would work with time series of imagery from multiple remote sensing sensor modalities. A benefit of our model is that it does not consider inter-year differences and thus is not affected by shifts in the color distributions of the imagery, but our experimental results do not explore the extent in which this is a useful property. Practically, there may be problems (such as shifts in geolocation accuracy) when applying TCM over stacks of imagery from different sources.

In summary, we hope TCM approach illustrated here will enable researchers to overcome the ``frozen'' labels of many emerging earth imagery datasets. Our lightweight approach to augment labels temporally should foster richer exploration of time series of satellite imagery and help us to understand the earth as it was, is, and will be.

\begin{acks}
We thank Microsoft Azure for support in cloud computing, and Schmidt Futures, Stanford Impact Labs, and the GRACE Communications Foundation for research support.
\end{acks}

\bibliographystyle{ACM-Reference-Format}
\bibliography{citations}


\begin{thebibliography}{39}


\ifx \showCODEN    \undefined \def \showCODEN     #1{\unskip}     \fi
\ifx \showDOI      \undefined \def \showDOI       #1{#1}\fi
\ifx \showISBNx    \undefined \def \showISBNx     #1{\unskip}     \fi
\ifx \showISBNxiii \undefined \def \showISBNxiii  #1{\unskip}     \fi
\ifx \showISSN     \undefined \def \showISSN      #1{\unskip}     \fi
\ifx \showLCCN     \undefined \def \showLCCN      #1{\unskip}     \fi
\ifx \shownote     \undefined \def \shownote      #1{#1}          \fi
\ifx \showarticletitle \undefined \def \showarticletitle #1{#1}   \fi
\ifx \showURL      \undefined \def \showURL       {\relax}        \fi
\providecommand\bibfield[2]{#2}
\providecommand\bibinfo[2]{#2}
\providecommand\natexlab[1]{#1}
\providecommand\showeprint[2][]{arXiv:#2}

\bibitem[\protect\citeauthoryear{Aminikhanghahi and Cook}{Aminikhanghahi and
  Cook}{2017}]%
        {aminikhanghahi2017survey}
\bibfield{author}{\bibinfo{person}{Samaneh Aminikhanghahi} {and}
  \bibinfo{person}{Diane~J Cook}.} \bibinfo{year}{2017}\natexlab{}.
\newblock \showarticletitle{A survey of methods for time series change point
  detection}.
\newblock \bibinfo{journal}{\emph{Knowledge and information systems}}
  \bibinfo{volume}{51}, \bibinfo{number}{2} (\bibinfo{year}{2017}),
  \bibinfo{pages}{339--367}.
\newblock


\bibitem[\protect\citeauthoryear{Anomaly}{Anomaly}{2015}]%
        {cafo-antibiotics}
\bibfield{author}{\bibinfo{person}{Jonathan Anomaly}.}
  \bibinfo{year}{2015}\natexlab{}.
\newblock \showarticletitle{What's Wrong with Factory Farming?}
\newblock \bibinfo{journal}{\emph{Public Health Ethics}}
  (\bibinfo{year}{2015}).
\newblock
\urldef\tempurl%
\url{https://ssrn.com/abstract=2392453}
\showURL{%
\tempurl}


\bibitem[\protect\citeauthoryear{Chen, Ouyang, and Agam}{Chen
  et~al\mbox{.}}{2019}]%
        {chen2019changenet}
\bibfield{author}{\bibinfo{person}{Ying Chen}, \bibinfo{person}{Xu Ouyang},
  {and} \bibinfo{person}{Gady Agam}.} \bibinfo{year}{2019}\natexlab{}.
\newblock \showarticletitle{ChangeNet: Learning to detect changes in satellite
  images}. In \bibinfo{booktitle}{\emph{Proceedings of the 3rd ACM SIGSPATIAL
  International Workshop on AI for Geographic Knowledge Discovery}}.
  \bibinfo{pages}{24--31}.
\newblock


\bibitem[\protect\citeauthoryear{Cubuk, Zoph, Shlens, and Le}{Cubuk
  et~al\mbox{.}}{2020}]%
        {cubuk2020randaugment}
\bibfield{author}{\bibinfo{person}{Ekin~D Cubuk}, \bibinfo{person}{Barret
  Zoph}, \bibinfo{person}{Jonathon Shlens}, {and} \bibinfo{person}{Quoc~V Le}.}
  \bibinfo{year}{2020}\natexlab{}.
\newblock \showarticletitle{Randaugment: Practical automated data augmentation
  with a reduced search space}. In \bibinfo{booktitle}{\emph{Proceedings of the
  IEEE/CVF Conference on Computer Vision and Pattern Recognition Workshops}}.
  \bibinfo{pages}{702--703}.
\newblock


\bibitem[\protect\citeauthoryear{Demir, Koperski, Lindenbaum, Pang, Huang,
  Basu, Hughes, Tuia, and Raskar}{Demir et~al\mbox{.}}{2018}]%
        {demir2018deepglobe}
\bibfield{author}{\bibinfo{person}{Ilke Demir}, \bibinfo{person}{Krzysztof
  Koperski}, \bibinfo{person}{David Lindenbaum}, \bibinfo{person}{Guan Pang},
  \bibinfo{person}{Jing Huang}, \bibinfo{person}{Saikat Basu},
  \bibinfo{person}{Forest Hughes}, \bibinfo{person}{Devis Tuia}, {and}
  \bibinfo{person}{Ramesh Raskar}.} \bibinfo{year}{2018}\natexlab{}.
\newblock \showarticletitle{Deepglobe 2018: A challenge to parse the earth
  through satellite images}. In \bibinfo{booktitle}{\emph{Proceedings of the
  IEEE Conference on Computer Vision and Pattern Recognition Workshops}}.
  \bibinfo{pages}{172--181}.
\newblock


\bibitem[\protect\citeauthoryear{Etten, Lindenbaum, and Bacastow}{Etten
  et~al\mbox{.}}{2018}]%
        {DBLP:journals/corr/abs-1807-01232}
\bibfield{author}{\bibinfo{person}{Adam~Van Etten}, \bibinfo{person}{Dave
  Lindenbaum}, {and} \bibinfo{person}{Todd~M. Bacastow}.}
  \bibinfo{year}{2018}\natexlab{}.
\newblock \showarticletitle{SpaceNet: {A} Remote Sensing Dataset and Challenge
  Series}.
\newblock \bibinfo{journal}{\emph{CoRR}}  \bibinfo{volume}{abs/1807.01232}
  (\bibinfo{year}{2018}).
\newblock
\showeprint[arxiv]{1807.01232}
\urldef\tempurl%
\url{http://arxiv.org/abs/1807.01232}
\showURL{%
\tempurl}


\bibitem[\protect\citeauthoryear{Evans and Malcom}{Evans and Malcom}{2020}]%
        {evans2020automated}
\bibfield{author}{\bibinfo{person}{Michael~J Evans} {and}
  \bibinfo{person}{Jacob~W Malcom}.} \bibinfo{year}{2020}\natexlab{}.
\newblock \showarticletitle{Automated Change Detection Methods for Satellite
  Data that can Improve Conservation Implementation}.
\newblock \bibinfo{journal}{\emph{bioRxiv}} (\bibinfo{year}{2020}),
  \bibinfo{pages}{611459}.
\newblock


\bibitem[\protect\citeauthoryear{Frangoul}{Frangoul}{2020}]%
        {frangoul}
\bibfield{author}{\bibinfo{person}{Anmar Frangoul}.}
  \bibinfo{year}{2020}\natexlab{}.
\newblock \showarticletitle{India has some huge renewable energy goals. But can
  they be achieved?}
\newblock \bibinfo{journal}{\emph{CNBC}} (\bibinfo{year}{2020}).
\newblock
\urldef\tempurl%
\url{https://www.cnbc.com/2020/03/03/india-has-some-huge-renewable-energy-goals-but-can-they-be-achieved.html}
\showURL{%
\tempurl}


\bibitem[\protect\citeauthoryear{Gupta, Welburn, Watson, and Yin}{Gupta
  et~al\mbox{.}}{2019}]%
        {gupta2019cnn}
\bibfield{author}{\bibinfo{person}{Ananya Gupta}, \bibinfo{person}{Elisabeth
  Welburn}, \bibinfo{person}{Simon Watson}, {and} \bibinfo{person}{Hujun Yin}.}
  \bibinfo{year}{2019}\natexlab{}.
\newblock \showarticletitle{CNN-Based Semantic Change Detection in Satellite
  Imagery}. In \bibinfo{booktitle}{\emph{International Conference on Artificial
  Neural Networks}}. Springer, \bibinfo{pages}{669--684}.
\newblock


\bibitem[\protect\citeauthoryear{Gupta and Shah}{Gupta and Shah}{2020}]%
        {gupta2020rescuenet}
\bibfield{author}{\bibinfo{person}{Rohit Gupta} {and} \bibinfo{person}{Mubarak
  Shah}.} \bibinfo{year}{2020}\natexlab{}.
\newblock \showarticletitle{Rescuenet: Joint building segmentation and damage
  assessment from satellite imagery}.
\newblock \bibinfo{journal}{\emph{arXiv preprint arXiv:2004.07312}}
  (\bibinfo{year}{2020}).
\newblock


\bibitem[\protect\citeauthoryear{Handan-Nader and Ho}{Handan-Nader and
  Ho}{2019}]%
        {handan2019deep}
\bibfield{author}{\bibinfo{person}{Cassandra Handan-Nader} {and}
  \bibinfo{person}{Daniel~E Ho}.} \bibinfo{year}{2019}\natexlab{}.
\newblock \showarticletitle{Deep learning to map concentrated animal feeding
  operations}.
\newblock \bibinfo{journal}{\emph{Nature Sustainability}} \bibinfo{volume}{2},
  \bibinfo{number}{4} (\bibinfo{year}{2019}), \bibinfo{pages}{298--306}.
\newblock


\bibitem[\protect\citeauthoryear{Handan-Nader, Ho, and Liu}{Handan-Nader
  et~al\mbox{.}}{2020}]%
        {handan2020deep}
\bibfield{author}{\bibinfo{person}{Cassandra Handan-Nader},
  \bibinfo{person}{Daniel~E Ho}, {and} \bibinfo{person}{Larry~Y Liu}.}
  \bibinfo{year}{2020}\natexlab{}.
\newblock \showarticletitle{Deep Learning with Satellite Imagery to Enhance
  Environmental Enforcement}.
\newblock \bibinfo{journal}{\emph{Data-Driven Insights and Decisions: A
  Sustainability Perspective. Elsevier}} (\bibinfo{year}{2020}).
\newblock


\bibitem[\protect\citeauthoryear{Hendrycks, Mu, Cubuk, Zoph, Gilmer, and
  Lakshminarayanan}{Hendrycks et~al\mbox{.}}{2019}]%
        {hendrycks2019augmix}
\bibfield{author}{\bibinfo{person}{Dan Hendrycks}, \bibinfo{person}{Norman Mu},
  \bibinfo{person}{Ekin~D Cubuk}, \bibinfo{person}{Barret Zoph},
  \bibinfo{person}{Justin Gilmer}, {and} \bibinfo{person}{Balaji
  Lakshminarayanan}.} \bibinfo{year}{2019}\natexlab{}.
\newblock \showarticletitle{Augmix: A simple data processing method to improve
  robustness and uncertainty}.
\newblock \bibinfo{journal}{\emph{arXiv preprint arXiv:1912.02781}}
  (\bibinfo{year}{2019}).
\newblock


\bibitem[\protect\citeauthoryear{Huang, Cao, and Li}{Huang
  et~al\mbox{.}}{2020}]%
        {huang2020automatic}
\bibfield{author}{\bibinfo{person}{Xin Huang}, \bibinfo{person}{Yinxia Cao},
  {and} \bibinfo{person}{Jiayi Li}.} \bibinfo{year}{2020}\natexlab{}.
\newblock \showarticletitle{An automatic change detection method for monitoring
  newly constructed building areas using time-series multi-view high-resolution
  optical satellite images}.
\newblock \bibinfo{journal}{\emph{Remote Sensing of Environment}}
  \bibinfo{volume}{244} (\bibinfo{year}{2020}), \bibinfo{pages}{111802}.
\newblock


\bibitem[\protect\citeauthoryear{Jung}{Jung}{2004}]%
        {jung2004detecting}
\bibfield{author}{\bibinfo{person}{Franck Jung}.}
  \bibinfo{year}{2004}\natexlab{}.
\newblock \showarticletitle{Detecting building changes from multitemporal
  aerial stereopairs}.
\newblock \bibinfo{journal}{\emph{ISPRS Journal of Photogrammetry and Remote
  Sensing}} \bibinfo{volume}{58}, \bibinfo{number}{3-4} (\bibinfo{year}{2004}),
  \bibinfo{pages}{187--201}.
\newblock


\bibitem[\protect\citeauthoryear{Listner and Niemeyer}{Listner and
  Niemeyer}{2011}]%
        {listner2011recent}
\bibfield{author}{\bibinfo{person}{Clemens Listner} {and}
  \bibinfo{person}{Irmgard Niemeyer}.} \bibinfo{year}{2011}\natexlab{}.
\newblock \showarticletitle{Recent advances in object-based change detection}.
  In \bibinfo{booktitle}{\emph{2011 IEEE International Geoscience and Remote
  Sensing Symposium}}. IEEE, \bibinfo{pages}{110--113}.
\newblock


\bibitem[\protect\citeauthoryear{Malpica, Alonso, Pap{\'\i}, Arozarena, and
  Mart{\'\i}nez De~Agirre}{Malpica et~al\mbox{.}}{2013}]%
        {malpica2013change}
\bibfield{author}{\bibinfo{person}{Jos{\'e}~A Malpica},
  \bibinfo{person}{Mar{\'\i}a~C Alonso}, \bibinfo{person}{Francisco Pap{\'\i}},
  \bibinfo{person}{Antonio Arozarena}, {and} \bibinfo{person}{Alex
  Mart{\'\i}nez De~Agirre}.} \bibinfo{year}{2013}\natexlab{}.
\newblock \showarticletitle{Change detection of buildings from satellite
  imagery and lidar data}.
\newblock \bibinfo{journal}{\emph{International Journal of Remote Sensing}}
  \bibinfo{volume}{34}, \bibinfo{number}{5} (\bibinfo{year}{2013}),
  \bibinfo{pages}{1652--1675}.
\newblock


\bibitem[\protect\citeauthoryear{Marcos, Hamid, and Tuia}{Marcos
  et~al\mbox{.}}{2016}]%
        {marcos2016geospatial}
\bibfield{author}{\bibinfo{person}{Diego Marcos}, \bibinfo{person}{Raffay
  Hamid}, {and} \bibinfo{person}{Devis Tuia}.} \bibinfo{year}{2016}\natexlab{}.
\newblock \showarticletitle{Geospatial correspondences for multimodal
  registration}. In \bibinfo{booktitle}{\emph{Proceedings of the IEEE
  Conference on Computer Vision and Pattern Recognition}}.
  \bibinfo{pages}{5091--5100}.
\newblock


\bibitem[\protect\citeauthoryear{Matsuoka and Yamazaki}{Matsuoka and
  Yamazaki}{2004}]%
        {matsuoka2004use}
\bibfield{author}{\bibinfo{person}{Masashi Matsuoka} {and}
  \bibinfo{person}{Fumio Yamazaki}.} \bibinfo{year}{2004}\natexlab{}.
\newblock \showarticletitle{Use of satellite SAR intensity imagery for
  detecting building areas damaged due to earthquakes}.
\newblock \bibinfo{journal}{\emph{Earthquake Spectra}} \bibinfo{volume}{20},
  \bibinfo{number}{3} (\bibinfo{year}{2004}), \bibinfo{pages}{975--994}.
\newblock


\bibitem[\protect\citeauthoryear{Moltzau}{Moltzau}{2020}]%
        {moltzau}
\bibfield{author}{\bibinfo{person}{Alex Moltzau}.}
  \bibinfo{year}{2020}\natexlab{}.
\newblock \showarticletitle{Estonia’s National Strategy for Artificial
  Intelligence}.
\newblock \bibinfo{journal}{\emph{Medium}} (\bibinfo{year}{2020}).
\newblock
\urldef\tempurl%
\url{https://medium.com/swlh/estonias-national-strategy-for-artificial-intelligence-2623259ddf4c}
\showURL{%
\tempurl}


\bibitem[\protect\citeauthoryear{Moynihan}{Moynihan}{2016}]%
        {Moynihan}
\bibfield{author}{\bibinfo{person}{Sheila Moynihan}.}
  \bibinfo{year}{2016}\natexlab{}.
\newblock \showarticletitle{Mapping Solar Potential in India}.
\newblock \bibinfo{journal}{\emph{{Office of Energy Efficiency and Renewable
  Energy}}} (\bibinfo{year}{2016}).
\newblock
\urldef\tempurl%
\url{https://www.energy.gov/eere/articles/mapping-solar-potential-india}
\showURL{%
\tempurl}


\bibitem[\protect\citeauthoryear{Osterberg and Wallinga}{Osterberg and
  Wallinga}{2004}]%
        {doi:10.2105/AJPH.94.10.1703}
\bibfield{author}{\bibinfo{person}{David Osterberg} {and}
  \bibinfo{person}{David Wallinga}.} \bibinfo{year}{2004}\natexlab{}.
\newblock \showarticletitle{Addressing Externalities From Swine Production to
  Reduce Public Health and Environmental Impacts}.
\newblock \bibinfo{journal}{\emph{American Journal of Public Health}}
  \bibinfo{volume}{94}, \bibinfo{number}{10} (\bibinfo{year}{2004}),
  \bibinfo{pages}{1703--1708}.
\newblock
\urldef\tempurl%
\url{https://doi.org/10.2105/AJPH.94.10.1703}
\showDOI{\tempurl}
\showeprint{https://doi.org/10.2105/AJPH.94.10.1703}
\newblock
\shownote{PMID: 15451736.}


\bibitem[\protect\citeauthoryear{Purdy}{Purdy}{2010}]%
        {purdy2010using}
\bibfield{author}{\bibinfo{person}{Ray Purdy}.}
  \bibinfo{year}{2010}\natexlab{}.
\newblock \showarticletitle{Using Earth observation technologies for better
  regulatory compliance and enforcement of environmental laws}.
\newblock \bibinfo{journal}{\emph{Journal of Environmental Law}}
  \bibinfo{volume}{22}, \bibinfo{number}{1} (\bibinfo{year}{2010}),
  \bibinfo{pages}{59--87}.
\newblock


\bibitem[\protect\citeauthoryear{Ronneberger, Fischer, and Brox}{Ronneberger
  et~al\mbox{.}}{2015}]%
        {ronneberger2015u}
\bibfield{author}{\bibinfo{person}{Olaf Ronneberger}, \bibinfo{person}{Philipp
  Fischer}, {and} \bibinfo{person}{Thomas Brox}.}
  \bibinfo{year}{2015}\natexlab{}.
\newblock \showarticletitle{U-net: Convolutional networks for biomedical image
  segmentation}. In \bibinfo{booktitle}{\emph{International Conference on
  Medical image computing and computer-assisted intervention}}. Springer,
  \bibinfo{pages}{234--241}.
\newblock


\bibitem[\protect\citeauthoryear{Roscher, Volpi, Mallet, Drees, and
  Wegner}{Roscher et~al\mbox{.}}{2020}]%
        {roscher2020semcity}
\bibfield{author}{\bibinfo{person}{Ribana Roscher}, \bibinfo{person}{Michele
  Volpi}, \bibinfo{person}{Cl{\'e}ment Mallet}, \bibinfo{person}{Lukas Drees},
  {and} \bibinfo{person}{Jan~Dirk Wegner}.} \bibinfo{year}{2020}\natexlab{}.
\newblock \showarticletitle{SemCity Toulouse: A benchmark for building instance
  segmentation in satellite images}.
\newblock \bibinfo{journal}{\emph{ISPRS Annals of Photogrammetry, Remote
  Sensing and Spatial Information Sciences}}  \bibinfo{volume}{5}
  (\bibinfo{year}{2020}), \bibinfo{pages}{109--116}.
\newblock


\bibitem[\protect\citeauthoryear{Soroka and Duren}{Soroka and Duren}{2020}]%
        {sorokaDataset}
\bibfield{author}{\bibinfo{person}{A.M. Soroka} {and} \bibinfo{person}{Z.
  Duren}.} \bibinfo{year}{2020}\natexlab{}.
\newblock \bibinfo{title}{{Poultry feeding operations on the Delaware,
  Maryland, and Virginia Peninsula from 2016 to 2017: U.S. Geological Survey
  data release}}.
\newblock \bibinfo{howpublished}{\url{https://doi.org/10.5066/P9MO25Z7}}.
\newblock


\bibitem[\protect\citeauthoryear{Straub, Koontz, and Loomis}{Straub
  et~al\mbox{.}}{2019}]%
        {Straub2019}
\bibfield{author}{\bibinfo{person}{Crista~L. Straub},
  \bibinfo{person}{Stephen~R. Koontz}, {and} \bibinfo{person}{John~B. Loomis}.}
  \bibinfo{year}{2019}\natexlab{}.
\newblock \bibinfo{booktitle}{\emph{Economic valuation of landsat imagery}}.
\newblock \bibinfo{type}{{T}echnical {R}eport}. \bibinfo{institution}{U.S.
  Geological Survey}, \bibinfo{address}{Reston, VA}.
\newblock
\showISSN{2019-1112}
\urldef\tempurl%
\url{https://doi.org/10.3133/ofr20191112}
\showDOI{\tempurl}
\newblock
\shownote{Report.}


\bibitem[\protect\citeauthoryear{Sun, Tzeng, Darrell, and Efros}{Sun
  et~al\mbox{.}}{2019}]%
        {sun2019unsupervised}
\bibfield{author}{\bibinfo{person}{Yu Sun}, \bibinfo{person}{Eric Tzeng},
  \bibinfo{person}{Trevor Darrell}, {and} \bibinfo{person}{Alexei~A Efros}.}
  \bibinfo{year}{2019}\natexlab{}.
\newblock \showarticletitle{Unsupervised domain adaptation through
  self-supervision}.
\newblock \bibinfo{journal}{\emph{arXiv preprint arXiv:1909.11825}}
  (\bibinfo{year}{2019}).
\newblock


\bibitem[\protect\citeauthoryear{Tewkesbury}{Tewkesbury}{2011}]%
        {tewkesbury2011mapping}
\bibfield{author}{\bibinfo{person}{A Tewkesbury}.}
  \bibinfo{year}{2011}\natexlab{}.
\newblock \showarticletitle{Mapping the extent of urban creep in Exeter using
  OBIA}. In \bibinfo{booktitle}{\emph{Proceedings of RSPSoc Annual
  Conference}}. \bibinfo{pages}{163}.
\newblock


\bibitem[\protect\citeauthoryear{Tewkesbury, Comber, Tate, Lamb, and
  Fisher}{Tewkesbury et~al\mbox{.}}{2015}]%
        {tewkesbury2015critical}
\bibfield{author}{\bibinfo{person}{Andrew~P Tewkesbury},
  \bibinfo{person}{Alexis~J Comber}, \bibinfo{person}{Nicholas~J Tate},
  \bibinfo{person}{Alistair Lamb}, {and} \bibinfo{person}{Peter~F Fisher}.}
  \bibinfo{year}{2015}\natexlab{}.
\newblock \showarticletitle{A critical synthesis of remotely sensed optical
  image change detection techniques}.
\newblock \bibinfo{journal}{\emph{Remote Sensing of Environment}}
  \bibinfo{volume}{160} (\bibinfo{year}{2015}), \bibinfo{pages}{1--14}.
\newblock


\bibitem[\protect\citeauthoryear{Tuia, Persello, and Bruzzone}{Tuia
  et~al\mbox{.}}{2016}]%
        {tuia2016domain}
\bibfield{author}{\bibinfo{person}{Devis Tuia}, \bibinfo{person}{Claudio
  Persello}, {and} \bibinfo{person}{Lorenzo Bruzzone}.}
  \bibinfo{year}{2016}\natexlab{}.
\newblock \showarticletitle{Domain adaptation for the classification of remote
  sensing data: An overview of recent advances}.
\newblock \bibinfo{journal}{\emph{IEEE geoscience and remote sensing magazine}}
  \bibinfo{volume}{4}, \bibinfo{number}{2} (\bibinfo{year}{2016}),
  \bibinfo{pages}{41--57}.
\newblock


\bibitem[\protect\citeauthoryear{Van~Etten and Hogan}{Van~Etten and
  Hogan}{2021}]%
        {van2021spacenet}
\bibfield{author}{\bibinfo{person}{Adam Van~Etten} {and}
  \bibinfo{person}{Daniel Hogan}.} \bibinfo{year}{2021}\natexlab{}.
\newblock \showarticletitle{The SpaceNet Multi-Temporal Urban Development
  Challenge}.
\newblock \bibinfo{journal}{\emph{arXiv preprint arXiv:2102.11958}}
  (\bibinfo{year}{2021}).
\newblock


\bibitem[\protect\citeauthoryear{Van~Etten, Hogan, Martinez-Manso, Shermeyer,
  Weir, and Lewis}{Van~Etten et~al\mbox{.}}{2021}]%
        {van2021multi}
\bibfield{author}{\bibinfo{person}{Adam Van~Etten}, \bibinfo{person}{Daniel
  Hogan}, \bibinfo{person}{Jesus Martinez-Manso}, \bibinfo{person}{Jacob
  Shermeyer}, \bibinfo{person}{Nicholas Weir}, {and} \bibinfo{person}{Ryan
  Lewis}.} \bibinfo{year}{2021}\natexlab{}.
\newblock \showarticletitle{The Multi-Temporal Urban Development SpaceNet
  Dataset}.
\newblock \bibinfo{journal}{\emph{arXiv preprint arXiv:2102.04420}}
  (\bibinfo{year}{2021}).
\newblock


\bibitem[\protect\citeauthoryear{Verbesselt, Hyndman, Newnham, and
  Culvenor}{Verbesselt et~al\mbox{.}}{2010}]%
        {verbesselt2010detecting}
\bibfield{author}{\bibinfo{person}{Jan Verbesselt}, \bibinfo{person}{Rob
  Hyndman}, \bibinfo{person}{Glenn Newnham}, {and} \bibinfo{person}{Darius
  Culvenor}.} \bibinfo{year}{2010}\natexlab{}.
\newblock \showarticletitle{Detecting trend and seasonal changes in satellite
  image time series}.
\newblock \bibinfo{journal}{\emph{Remote sensing of Environment}}
  \bibinfo{volume}{114}, \bibinfo{number}{1} (\bibinfo{year}{2010}),
  \bibinfo{pages}{106--115}.
\newblock


\bibitem[\protect\citeauthoryear{Wang, Li, Ying, Cheng, Wang, Li, Hu, Liang,
  Yu, Huang, et~al\mbox{.}}{Wang et~al\mbox{.}}{2012}]%
        {wang2012china}
\bibfield{author}{\bibinfo{person}{Lei Wang}, \bibinfo{person}{Congcong Li},
  \bibinfo{person}{Qing Ying}, \bibinfo{person}{Xiao Cheng},
  \bibinfo{person}{Xiaoyi Wang}, \bibinfo{person}{Xueyan Li},
  \bibinfo{person}{Luanyun Hu}, \bibinfo{person}{Lu Liang}, \bibinfo{person}{Le
  Yu}, \bibinfo{person}{HuaBing Huang}, {et~al\mbox{.}}}
  \bibinfo{year}{2012}\natexlab{}.
\newblock \showarticletitle{China’s urban expansion from 1990 to 2010
  determined with satellite remote sensing}.
\newblock \bibinfo{journal}{\emph{Chinese Science Bulletin}}
  \bibinfo{volume}{57}, \bibinfo{number}{22} (\bibinfo{year}{2012}),
  \bibinfo{pages}{2802--2812}.
\newblock


\bibitem[\protect\citeauthoryear{Weir, Lindenbaum, Bastidas, Etten, McPherson,
  Shermeyer, Kumar, and Tang}{Weir et~al\mbox{.}}{2019}]%
        {weir2019spacenet}
\bibfield{author}{\bibinfo{person}{Nicholas Weir}, \bibinfo{person}{David
  Lindenbaum}, \bibinfo{person}{Alexei Bastidas}, \bibinfo{person}{Adam~Van
  Etten}, \bibinfo{person}{Sean McPherson}, \bibinfo{person}{Jacob Shermeyer},
  \bibinfo{person}{Varun Kumar}, {and} \bibinfo{person}{Hanlin Tang}.}
  \bibinfo{year}{2019}\natexlab{}.
\newblock \showarticletitle{Spacenet MVOI: a multi-view overhead imagery
  dataset}. In \bibinfo{booktitle}{\emph{Proceedings of the IEEE/CVF
  International Conference on Computer Vision}}. \bibinfo{pages}{992--1001}.
\newblock


\bibitem[\protect\citeauthoryear{Yang}{Yang}{2018}]%
        {yang}
\bibfield{author}{\bibinfo{person}{Siyu Yang}.}
  \bibinfo{year}{2018}\natexlab{}.
\newblock \showarticletitle{How to extract building footprints from satellite
  images using deep learning}.
\newblock \bibinfo{journal}{\emph{Microsoft}} (\bibinfo{year}{2018}).
\newblock
\urldef\tempurl%
\url{https://azure.microsoft.com/en-us/blog/how-to-extract-building-footprints-from-satellite-images-using-deep-learning/}
\showURL{%
\tempurl}


\bibitem[\protect\citeauthoryear{Zhang, Liu, Gros, and Tiecke}{Zhang
  et~al\mbox{.}}{2017}]%
        {zhang2017building}
\bibfield{author}{\bibinfo{person}{Amy Zhang}, \bibinfo{person}{Xianming Liu},
  \bibinfo{person}{Andreas Gros}, {and} \bibinfo{person}{Tobias Tiecke}.}
  \bibinfo{year}{2017}\natexlab{}.
\newblock \showarticletitle{Building detection from satellite images on a
  global scale}.
\newblock \bibinfo{journal}{\emph{arXiv preprint arXiv:1707.08952}}
  (\bibinfo{year}{2017}).
\newblock


\bibitem[\protect\citeauthoryear{Zhao, Kang, Jung, and Sohn}{Zhao
  et~al\mbox{.}}{2018}]%
        {zhao2018building}
\bibfield{author}{\bibinfo{person}{Kang Zhao}, \bibinfo{person}{Jungwon Kang},
  \bibinfo{person}{Jaewook Jung}, {and} \bibinfo{person}{Gunho Sohn}.}
  \bibinfo{year}{2018}\natexlab{}.
\newblock \showarticletitle{Building extraction from satellite images using
  mask R-CNN with building boundary regularization}. In
  \bibinfo{booktitle}{\emph{Proceedings of the IEEE Conference on Computer
  Vision and Pattern Recognition Workshops}}. \bibinfo{pages}{247--251}.
\newblock


\end{thebibliography}

\end{document}